\documentclass{article}

\usepackage[nonatbib]{neurips_2020}
\usepackage{caption}
\usepackage{subcaption}
\usepackage[utf8]{inputenc} % allow utf-8 input
\usepackage[T1]{fontenc}    % use 8-bit T1 fonts
\usepackage{hyperref}       % hyperlinks
\usepackage{url}            % simple URL typesetting
\usepackage{booktabs}       % professional-quality tables
\usepackage{amsfonts}       % blackboard math symbols
\usepackage{nicefrac}       % compact symbols for 1/2, etc.
\usepackage{microtype}      % microtypography
\usepackage{graphicx}
\usepackage{enumitem}
\usepackage{comment}
\usepackage{bbm}
\usepackage{amsmath,amssymb} % define this before the line numbering.
\usepackage{color}
\usepackage{multirow}
\DeclareMathOperator*{\argmin}{argmin}

\title{Injecting Knowledge in Data-driven \\ Vehicle Trajectory Predictors}
\author{Mohammadhossein Bahari\\
   \And
   Ismail Nejjar \\
    \And
    Alexandre Alahi \\
  \AND
  \texttt{Ecole polytechnique fédérale de lausanne (EPFL)} \\
  \texttt{mohammadhossein.bahari@epfl.ch} }

\begin{document}

\maketitle
\begin{abstract}
% there have been two almost seperate line of researchs
% knowledge-aware models which lead to realistic outputs,
% data-driven are able to capture non modeled knowledes + learning patterns leading to leading to better performance.
% We want to merge both worlds by knowledge predicting a base prediction and residual adds data-drived information
% We demonstrate the performance on a challenging real-world dataset
Vehicle trajectory prediction tasks have been commonly tackled from two distinct perspectives: either with knowledge-driven methods or more recently with data-driven ones. On the one hand, we can explicitly implement domain-knowledge or physical priors such as anticipating that vehicles will follow the middle of the roads. While this perspective leads to feasible outputs, it has limited performance due to the difficulty to hand-craft complex interactions in urban environments. On the other hand, recent works use data-driven approaches which can learn complex interactions from the data leading to superior performance. However, generalization, \textit{i.e.}, having accurate predictions on unseen data, is an issue leading to unrealistic outputs. In this paper, we propose to learn a "Realistic Residual Block" (RRB), which effectively connects these two perspectives. Our RRB takes any off-the-shelf knowledge-driven model and finds the required residuals to add to the knowledge-aware trajectory. Our proposed method outputs realistic predictions by confining the residual range and taking into account its uncertainty. We also constrain our output with Model Predictive Control (MPC) to satisfy kinematic constraints. Using a publicly available dataset, we show that our method outperforms previous works in terms of accuracy and generalization to new scenes. We will release our code and data split here: https://github.com/vita-epfl/RRB.

\end{abstract}

\section{Introduction}

While driving, humans have this powerful capability to anticipate other drivers’ decisions. Similarly, an autonomous vehicle should have the same prediction capability to safely navigate alongside human drivers. Some researchers addressed the vehicle trajectory prediction task, also known as microscopic traffic modeling, by building hand-crafted functions based on the available domain-knowledge to model average driving behaviors \cite{keyvan2016categorization,hondakf,leader-follower,coscia2016point,coscia2018long,xu2015asymmetric}. These methods are interpretable and usually lead to a set of feasible predictions. However, they have limited performance since they not only miss non-average behaviors, but also are not able to model complex interactions. Conversely, recent works solely rely on experience, \textit{i.e.}, learning from data, mostly using neural networks \cite{zhang2019simultaneous,xie2019data,social_lstm,alahi2017learning,kothari2020human,mult_future_pred}. Using large amount of data helps these methods to achieve accurate predictions without explicitly modeling the domain-knowledge. However, their predictions are not essentially realistic (on-road) and in some cases, even counter-intuitive.
Moreover, they are prone to overfitting on the training data or poor performance on out-of-distribution data. Combining the domain-knowledge and the data will benefit from the strengths of both approaches and avoid their shortcomings.

We argue that driving is a skill learned from domain-knowledge and experience. The former is typically driven by physical constraints such as respecting road constraints or avoiding collisions. The later is driven by social conventions \textit{e.g.}, the way drivers interact with each other or the safe way to enter a roundabout. Consequently, learning to predict vehicle trajectories can be re-framed as learning to combine both knowledge-driven and data-driven methods. 
Human's ability to employ both knowledge and data is not specific to driving. Researchers believe that humans learn rich representations (we refer to as knowledge) as well as patterns from observed examples in everyday life \cite{lake2017building}. This allows them to learn from fewer examples and to generalize to new conditions \cite{lake2015human}. 

%It is believed that human learn rich representations (which we refer to as knowledge) as well as patterns from observed examples \cite{lake2017building}. This allows them to learn from less examples and to generalize to new conditions \cite{lake2015human}. Similarly, people use both their world-knowledge and their experience while driving. A driver has some general knowledge about driving, for example the fact that he has to respect road constraints or avoid collisions. On the other hand, there are many skills need to be learned through experience e.g., the way drivers interact with each other or the way they enter a roundabout. Consequently, the two factors of knowledge and experience (data) are required for anticipating driving behavior. 

% talk about vehicle prediction, inputs and outputs, then previous works on each side (one or two), knowledge or experience, need to use both
% researchers tried to anticipate knowledge 

%talk about works tried to add constraints
A popular way of adding domain-knowledge to the neural network is by adding constraints to the problem and optimizing the network under these constraints.
Authors in \cite{uber-scene} proposed a road-loss which can be interpreted as an approximation of the scene constraint in order to avoid off-road predictions. However, directly optimizing the model under constraints makes the optimization difficult and leads to sub-optimal results \cite{posterior, weakly_supervised}. 

%talk about challenges , limitations of previous works,
% how to fuse knowledge with nn in a way that still the network can be trained using data (i.e ), not to lose added knowledge and realisticity (i.e. )
% {Not so sure(difficult to add multiple constraints)}, and formulate specially in a differentiable form, 
There exist two main challenges towards creating a knowledge-driven and data-driven model. First, the combined model should be differentiable so that the data-driven part can be trained. This means that the integration needs careful design as the knowledge-driven part of the model usually is not differentiable. 
%First, fusing knowledge into neural network requires differentiable knowledge-driven functions as the final model should be trainable. Consequently, designing the model structure is challenging as most of knowledge-driven functions are not differentiable. For example, formulating the scene knowledge or driving rules in a differentiable way is not trivial. 
%Similarly when adding constraint to neural networks, in many cases, the knowledge cannot be expressed as a differentiable function which prohibits using posterior regularization.
The second challenge is to preserve the benefits of both worlds after merging them \textit{i.e.}, the final output should be realistic despite the fact that neural network might generate unrealistic outputs.% and the added knowledge should allow the network to learn from data.

\begin{figure}[t]
\centering
\includegraphics[height=4.2cm]{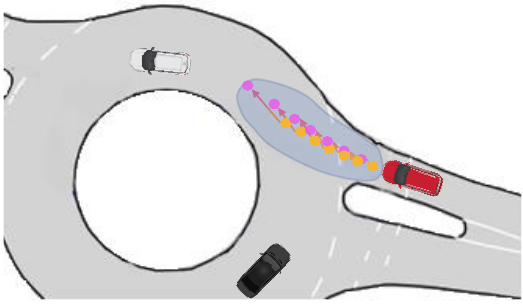}
\caption{Illustration of our Realistic Residual Block (RRB) model. The knowledge-based model generates a scene-compliant trajectory (yellow dots) which could not effectively account for other agents, hence, is too conservative. Our data-driven RRB improves the prediction (purple dots) by adding confined residuals conditioned on other agents in the scene (illustrated by the  arrows). The blue region shows the physically-constrained output space for our RRB predictions.}
\label{fig:pull}
\end{figure}

% proposed solution 
In this work, we address the aforementioned challenges. In the proposed approach shown in figure \ref{fig:pull}, the knowledge-driven (KD) trajectory is achieved by any knowledge-driven model. The KD output is then taken as input by our Realistic Residual Block (RRB) and the residuals required to be added to KD trajectory are found. In other words, the KD trajectory estimates coarse-grained behaviors based on the common driving performances while the residuals address fine-grained behaviors coming from non-modeled social interactions in KD trajectory as well as the long-tail of performances. This structure allows imposing knowledge by any function while allowing the residual block to be trainable. Moreover, in order not to diminish the feasibility of KD prediction, we can physically constrain the output of our RRB (\textit{i.e.}, the allowed feasible offset) and leverage its uncertainty in the combination. Hence, our RRB uses a physically-constrained Inverse-variance weighting approach to add feasible and confident residuals to the KD trajectory. We finally constrain the final output with Model Predictive Controller (MPC) to ensure kinematic-feasibility of predictions. %To the best of our knowledge, we are the first one that employ MPC in the vehicle trajectory prediction task.

% Moreover, the kinematic constraint should be added. previous works used kinematic layer, but it prevents... 

%RRB can bring domain-knowledge to the outputs, however, the outputs may be kinematically-infeasible. Authors in \cite{uber_kinematic,Conditional} tackled the problem by adding the bicycle model \cite{bicycle_model} as a layer to the model. This approach requires the outputs of the model to be control commands rather than coordinate values so that the bicycle model finds the kinematically-consistent trajectory. Despite the simplicity and effectiveness of this approach, having control commands as the outputs prevents any further knowledge injection to the model. Since any changes to the forecasted trajectory can lead to a non-kinematically-feasible trajectory. Moreover, the knowledge-based model and the residual block cannot estimate the output as control commands. In contrast to the previous works, we propose using a Model Predictive Controller (MPC) to satisfy kinematic constraints. This allows the outputs to be in the coordinate format, mitigating the mentioned problems. %To the best of our knowledge, it is the first time that MPC is used in the vehicle trajectory prediction task

The contributions of this work can be summarized as: (1) proposing a Realistic Residual Block (RRB) as an effective way of leveraging both knowledge and data in vehicle trajectory prediction. Our RRB complements the knowledge-driven output with realistic and confident outputs. (2) Using Model Predictive Controller (MPC) to bring kinematic constraints to the vehicle trajectory prediction task with latent control variables.
(3) Demonstrate the generalization of our approach to new scenes and the multimodal case. 
%\item Evaluating the proposed method on synthetic and real datasets and achieving state-of-the-art results

\section{Related Work}
{\bf Vehicle trajectory prediction:} Pioneering works addressed vehicle trajectory prediction problem by means of knowledge-driven methods.
Researchers in \cite{hondakf} used Kalman filter \cite{kalman} to predict vehicle future trajectory. In order to incorporate scene information, \cite{benz} proposed associating vehicle's positions with the lanes of the road. Vehicle-vehicle interaction is addressed in \cite{leader-follower} to predict the longitudinal motion of a target vehicle. In \cite{xu2015asymmetric}, an asymmetric optimal velocity model is presented to capture the asymmetry between acceleration and deceleration and \cite{keyvan2016categorization} studied lane change decision empirically. 
%Authors in \cite{coscia2018long} used knowledge-based circular distributions to model the impact of static contexts on human behavior. %They build circular distributions for past observation, semantic scene, motion model, and direction to the goal and find the final prediction by the product of the distributions.
% \cite{social-forces} proposed a hand-crafted model referred to as the Social-force model to capture social behaviors. They model interactions between pedestrians by means of social fields determined by repulsive and attractive forces. 
On the other hand, many researchers tackle the problem by leveraging data-driven models. A deep belief network is proposed in \cite{xie2019data} to model lane-changing behavior. Authors in \cite{zhang2019simultaneous,convolution-pool} model lane-changing and car following behaviors simultaneously by employing long short-term memory (LSTM) neural networks and convolutional social pooling respectively. Authors in \cite{carnet} used an attention module to incorporate scene features into an LSTM model. An inverse optimal control (IOC) ranking module is used in \cite{desire} to determine the most likely hypotheses incorporating scene context and interactions. Moreover, some researchers addressed the multimodal nature of human behavior prediction by using Winner-Takes-All (WTA) loss method \cite{uber_multimodal,makansi2019overcoming}.
While all mentioned works rely on knowledge or data exclusively, our solution benefits from both approaches.

%Interaction between agents was addressed by a data-driven model in  \cite{social_lstm}. They showed the complex interaction between agents could be learned more effectively than hand-crafted counterparts by sharing hidden states of Long-Short Term Memory (LSTM) networks.  

{\bf Injecting domain-knowledge to neural networks:}
% dk can be injected by two ways, inside model or learning procedure.
% model change: CNN,  , but are usually problem specific
% learning procedure: hard constraint, soft constraint, mainly poster regul, constrained cnn, teacher-student, learning constraints from data
% our proposed approach is model change but prob agnostic, doesn't need to be differentiable
Knowledge has been incorporated into the non-data-driven models to preserve realisticity and improve generalization. In \cite{an2015space,tang2016estimating}, space-time constraints were employed to confine the outputs to the feasible range.
Researchers have proposed different means for incorporating domain-knowledge into neural networks \cite{willard2020integrating, borghesi2020improving, survey_knowledge}. It can be injected into neural networks by designing specialized network structures \cite{francca2014fast,garcez2019neural}. As an example, Convolutional neural network (CNN) was created by changing multilayer perceptron (MLP) networks, considering image specifications \cite{cnn-lecun}. Another approach for fusing neural networks with domain knowledge is to modify the learning algorithm. This can be done by exposing knowledge-driven negative samples to the model \cite{liu2020social} or adding constraints to the outputs of the network \cite{survey_knowledge}. Other approaches are adding knowledge to the training data \cite{karpatne2017physics} and post-processing \cite{fang2017object}. From another perspective, some researchers used neural networks to address the imperfection of knowledge-driven models. They used knowledge-driven models to find biases of the data and the neural network compensates the error by outputting required residuals. \cite{zeng2020tossingbot, silver2018residual} predict residuals on top of a physics-based robotic controller and \cite{kani2017dr} finds residual minimiser of numerically discretized differential equations. 

A key limitation in residual modeling is by adding neural network-based residuals, the physics-based constraints, which are required for a realistic prediction, are hard to preserve \cite{willard2020integrating}. In this paper, we mitigate this limitation by confining the residual values. Moreover, we replaced naive addition of residuals by variance-based integration methods. 

In the context of vehicle trajectory prediction, previous works attempt to add scene knowledge by adding penalty terms to the loss function. Researchers in \cite{waymo} defined an on-road loss to keep the predictions inside the road. The proposed loss requires the output to be an occupancy heatmap, which is computationally expensive to achieve and also not compatible with most of the trajectory prediction works \cite{uber-scene}. Authors in \cite{uber-scene} proposed an off-road loss defined as the euclidean distance between each predicted waypoint and the nearest drivable point to penalize off-road predictions. Their experiments show that although off-road predictions are reduced, the performance deteriorates in terms of $\ell_2$ loss. This is due to the fact that direct optimization of the model with constraints which are non-linear with respect to model parameters is difficult and leads to sub-optimal solutions \cite{posterior, weakly_supervised}.
In this paper, we propose a new method for adding scene knowledge to the model. 
%By leveraging a high definition map, they generate a rasterized image that includes the drivable and non-drivable regions for each vehicle. Their first loss is a euclidean-loss between each predicted waypoint and the nearest drivable point.

%This is probably due to the fact that their off-road loss aims at correcting the direction of the prediction while at the same time manipulates the absolute velocities and misguides the network from the human-like behavior, e.g. the network might take the edge of the road as a goal and deviates an on-road prediction towards that. Our proposed solution is an end-to-end learning model, thus, although it brings the constraint of the road to the model, it only uses the imitation loss leading to a human-like behavior.

{\bf Kinematically-feasible predictions:}
Kinematic constraints are physical rules that need to be satisfied for a realistic vehicle prediction.
%For a vehicle prediction to be realistic, respecting both road constraints and kinematic constraints are essential. 
Authors in \cite{uber_kinematic} showed that the predictions of the neural network model were not essentially kinematically-possible. Researchers in \cite{uber_kinematic,Conditional} solved the problem by adding the bicycle model \cite{bicycle_model} as a layer to the model. 
The network estimates control commands instead of coordinates and the kinematic layer converts them to a feasible trajectory. Despite the effectiveness of this approach, in many cases, it cannot be employed as most of the off-the-shelf models predict coordinates rather than control commands. Moreover, having control commands as the outputs prevents any further knowledge injection to the model's output, since any changes to the predicted trajectory can lead to a non-kinematically-feasible trajectory. In contrast, we propose using a Model Predictive Controller (MPC) to satisfy kinematic constraints. This allows the outputs to be in the coordinate format while the control commands are latent variables and mitigates the mentioned problems. Model Predictive Controller is commonly used for planning \cite{benz,human-centered,learning-based,surround-vehicle,cooperation-aware}. In this work, we show how it can also be used for the vehicle trajectory prediction task.

\section{Proposed Method}
\begin{figure}[t]
\centering
\includegraphics[height=2.9cm]{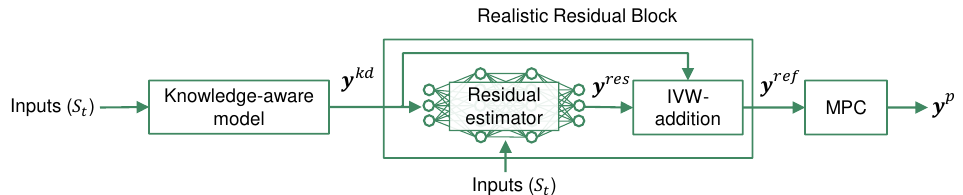}
\caption{Our proposed Realistic Residual Block (RRB) takes as input (i) the state of the scene, $S_t$, and (ii) the knowledge-driven prediction, $\mathbf{y}^{kd}$. The Residual estimator block builds physically-constrained residuals $\mathbf{y}^{res}$. Then, the IVW-addition block merges $\mathbf{y}^{kd}$ and $\mathbf{y}^{res}$ by Inverse-Variance Weighted sum and forms $\mathbf{y}^{ref}$ according to equation \ref{merge_equation}. Finally, Model Predictive Controller (MPC) satisfies the kinematic constraints.
}
\label{fig:model}
\end{figure}

Humans have a clear understanding of the domain-knowledge while driving \textit{e.g.}, where the drivable and non-drivable areas are. Moreover, they learn specific aspects of driving by experience \textit{e.g.}, interacting with other agents. However, benefiting from the domain-knowledge and learning from experience simultaneously is challenging. We address the problem by proposing a Realistic Residual Block (RRB). RRB finds data-driven residuals conditioned on the knowledge-aware prediction. The output is a physically-constrained Inverse-Variance Weighted (IVW) sum of the knowledge-driven (KD) trajectory with the residuals. Finally, Model Predictive Control (MPC) is incorporated to satisfy kinematic constraints. Figure \ref{fig:model} shows a high level picture of our proposed model. We will explain each part of the model in the following subsections.  
%The DNN predictor estimates the $\mathbf{y}_{i,t}^{nn}$ output. This output is mapped to the domain-knowledge using predefined mapping functions to find  $\mathbf{y}_{i,t}^{kd}$. This allows the model to learn complex functions directly from the data and still utilize the available domain-knowledge. For instance, the vehicle's velocity, which reflects driver's intention, can be found by the neural network and then constrained by the scene.
% However, despite the robustness and generalizability, the knowledge-driven output would lose accuracy as the prediction task has a bit of stochasticity which is difficult to be formulated in the mapping functions. We add  the residual attention block to cover this stochasticity. The residual attention block takes the knowledge-driven output as a prior and finds the residual required to be added to it based on the knowledge-driven output, neural network's output, and the context. Context can be any parts of the inputs which help to find better residuals. In our specific problem, a concatenation of encoded features used in the decoder is considered as the context.
\subsection{Problem Formulation}
The goal of our model is to predict future positions of a vehicle given its history and surroundings. Therefore, the state of the scene input to the model at time $t$, $S_t$, consists of an image of the scene %$I_t$ 
and histories of ego-vehicle and other vehicles in the scene. By ego-vehicle, we refer to the vehicle whose future is predicted. The observation vectors for ego-vehicle and other vehicles' history comprises of $T_{o}$ observed 2D coordinates of them in the world space. Without loss of generality, each time, one of the agents in the scene is taken as the ego-vehicle.
%, $H_t = [\mathbf{x}_{0,t}^{o}, \mathbf{x}_{1,t}^{o}, ..., \mathbf{x}_{K_t,t}^{o}]$, for $T_{o}$ observation frames where $\mathbf{x}_{i,t}^{o} =[(x_{i,t-T_{o}},y_{i,t-T_{o}}),...,(x_{i,t},y_{i,t})]$ is the observation vector of agent $i$ at time $t$, $(x,y)$ is the 2D coordinate in the pixel space or world space and $K_t$ is the number of tracked agents at time $t$. Without loss of generality, each time, one of the agents in the scene is taken as the primary agent and its future positions %denoted as $x = ...$ for x-positions of the agent $i$ and $y$ for y-positions 
% are predicted. Hence, we won't write the index $i$ in the next sections for simplicity. %It is worth mentioning that in order to remove the local information of the ego-vehicle, the inputs and outputs of the model are relative velocities.
At time $t$, the model predicts ${\mathrm{\mathbf{y}}^{ref}_{t}}$ for the next $T_{p}$ prediction frames as a Gaussian distribution with mean $\mu$ and variance $\sigma$ as:
\begin{gather}
    %p(z|s) = \sum_{m=1}^{M} \pi_m(s)\prod_{t=1}^{t_{pred}} \phi_m(z_t|s)
    % \mathbf{y}^{ref}_{i,t} = [\mathrm{y}^{ref}_{i,t+1},\mathrm{y}^{ref}_{i,t+2},...,\mathrm{y}^{ref}_{i,t+T_p}]
    p(\mathbf{y}^{ref}_{t}|S_t,\boldsymbol{\mu}_{t},\boldsymbol{\sigma}_{t}) = \prod_{j=1}^{T_{p}} \mathcal{N}( \mathrm{y}^{ref}_{t+j}|S_{t},\mu_{t+j},\sigma_{t+j}).
\end{gather}
Note that we indicate the sequence of values in a bold text. We omit the index $t$ in the rest of the paper for simplicity.

\subsection{Knowledge-aware Prediction} \label{knowledge-aware-section}
%There exist plenty of domain-knowledge that can be used in the model like scene knowledge
Our proposed method is flexible in the choice of the knowledge-aware model. Hence, any off-the-shelf knowledge-driven prediction model can be employed to create the KD trajectory. In section \ref{ablation-section}, we show the robustness of our RRB to different knowledge-driven models. In this work, we make the KD trajectory by utilizing the scene knowledge, the most influential domain-knowledge in vehicle prediction task. Inspired by \cite{benz}, we use the lanes of the road to form the scene-compliant trajectory $\mathbf{y}^{kd}$. To have a probabilistic framework, we take KD trajectory as the mean of a Gaussian distribution with fixed variances which are approximated by the statistics of the training data.
%We explored different strategies for estimating vehicle's speed and chose \cite{the-simpler-the-better} as the best solution.
%In order to create our KD prediction, first, we used Constant Velocity model \cite{the-simpler-the-better} to forecast the vehicle's future positions. Then, similar to \cite{benz}, we use the lanes of the road to perceive scene and map predicted points to the lanes of the road, leading to the scene-compliant trajectory $\mathbf{y}^{kd}$. 

%\textit{should be improved : It is desirable to have an RRB which is agnostic to the knowledge-aware model. It's shown in section \ref{ablation-section} that simply training the model does not lead to a robust RRB. Hence, inspired by \cite{dart} we expand the space of input by adding noise to the Constant Velocity predicted trajectory while training. This, noticeably improves the generalization power of RRB block approved by our ablation study in the next section. }

\subsection{Realistic Residual Block} \label{rrb-section}
% How knowledge is injected
%   we need a way to add the scene information
%   we give the net the flexiblity to deviate from rules
%    the attention module learns to ... . 
%    it learns to obey center line when it is not sure about the output
%   Although still a knowledge-driven, but the task is easier and generalizable
% Why MPC and how
\begin{figure}
    \centering
    \includegraphics[height=4.9cm]{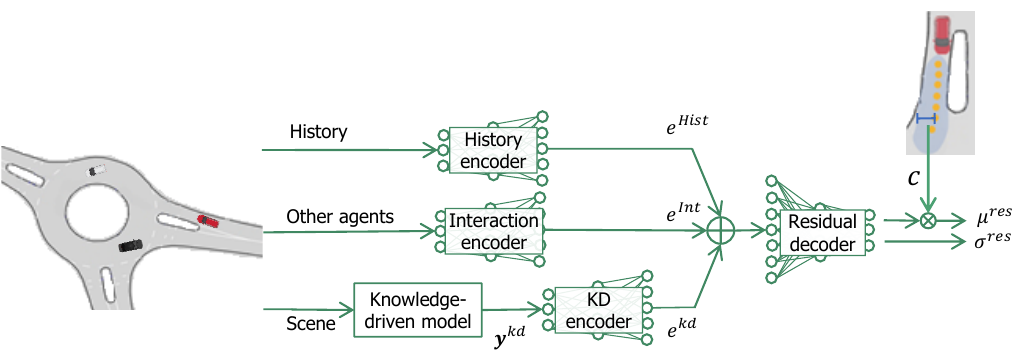}
    \caption{The 'Residual estimator' block inside our RRB. Three encoders are employed to embed the history, the interactions and the KD trajectory. The residual decoder integrates the features and estimates the residual distribution. The mean of the distribution is confined to $C$, a real-word extracted  parameter.}
    \label{fig:residual_estimator_model}
\end{figure}
%The KD trajectory addresses average behaviors. Our proposed data-driven Realistic Residual Block (RRB) learns the missing complex interaction.
While KD trajectory reflects average behaviors with respect to the scene, our proposed data-driven Realistic Residual Block learns the missing complex interactions. The architecture of RRB is depicted in Figure \ref{fig:residual_estimator_model}. %The residuals are learned by our careful designed Residual estimator. 
The learned residuals are first confined %by a real-word parameter 
and then merged with the KD prediction by the IVW-addition block. This will lead to feasible and confident predictions. Both Residual estimator and integration mechanism are explained in the following subsections.

\subsubsection{Residual Estimator}
Our Residual estimator's structure is shown in Figure \ref{fig:residual_estimator_model}. 
%The state input is processed by an encoder-decoder network.
%Similar to \cite{waymo}, the scene preprocessing block builds the normalized cropped segmented image of the scene around the vehicle at the prediction time (\textit{i.e.}, actor-centric representation). A multilayer CNN then extracts the scene's features. 
The inputs are the KD trajectory and $S_t$ consisting of the image of the scene and history of agents in coordinate format. The history of ego-vehicle is processed by the history encoder to find the driver's intention leading to the feature $\mathbf{e}^{Hist}$. We model the interaction among agents in a simple yet effective approach. Inspired by how vehicles interact in the real-word, our preprocessing eliminates non-interacting agents in the scene. In contrast with most previous works \cite{social_lstm, mult_future_pred} which consider surrounding agents, it excludes all the agents behind the vehicle as usually, vehicles interact with the agents in front. Moreover, as only the closest vehicles can impact the driver's behavior, it only preserves a set of closest vehicles and passes them to the interaction encoder model to get the interaction features $\mathbf{e}^{Int}$. Our experiments show the effectiveness of the model in learning interactions between agents.
The KD trajectory $\mathbf{y}^{kd}$, is encoded into the vector $\mathbf{e}^{kd}$ by the KD encoder network. All the encoders are feed-forward networks with ReLU non-linearities.
% according to the following equation:
% \begin{equation}
% \begin{split}
    % \mathbf{e}^{kd} = \phi(\mathbf{y}^{kd};W^{kd}), \hspace{2mm}
    % \mathbf{e}^{I} = \phi(I;W^{I}),
% \end{split}    
% \end{equation}
% where $\phi(.)$ is an embedding function with ReLU non-linearity and $W^{kd}$
% is embedding weights.
Finally, the residual decoder estimates the residuals given the concatenated features as:
\begin{equation}
\begin{split}
     (\hat{\mu}^{res}, \sigma^{res}) &= \mathit{MLP}([\mathbf{e}^{Hist},\mathbf{e}^{Int},\mathbf{e}^{kd}]; W^{res}),\\
     \mu^{res} &= { C}\hat{\mu}^{res},\\
\end{split}
\end{equation}  
where $(\mu^{res}, \sigma^{res})$ are parameters of the Gaussian distribution $\mathbf{y}^{res}$ and $\mathit{MLP}$ is a multilayer perceptron network parameterized by $W^{res}$. 
% with Tanh and Sigmoid non-linearities for $\hat{\mu}^{res}$ and $\sigma^{res}$ respectively.
Note that we bound the range of $\hat{\mu}^{res}$ to $(-1,1)$ by using $\tanh$ activation function in the last layer. $\mu^{res}$ is the scaled version of $\hat{\mu}^{res}$ by $ C$ to adjust the max feasible deviation from the center of the road. $ C$ is a real-word extracted parameter which can be adapted to each scene. In our experiments, we chose $C$ equal to the half of the minimum width of the road in each scene. 

\subsubsection{IVW-addition}
%other names for IVW:  uncertainty-aware averaging (UAA) or least-variance averaging, precision-weighted" average (according to wikipedia), vrience-optimal weighting
The merged trajectory $\mathbf{y}^{ref}$ can be achieved by simply adding residual Gaussians to the KD Gaussian prediction, referred to "A-RRB" baseline in Section \ref{metrics-section}. However, adding uncertain residuals to the KD trajectory can impair the KD predictions. To mitigate that, we utilize Inverse-Variance Weighting (IVW) \cite{ivw} to scale residuals according to their uncertainties which results in the most certain output (in terms of output variance) among all weighted averages \cite{hartung2011statistical}. 
%The parameters of the final reference trajectory is achieved by the following equation. We demonstrate the proof of the equation in Appendix \ref{appendix.a}. 
% \begin{equation}
%  \begin{split}
%     %\mu^{ref} = {\mu^{res}*C/{\sigma^{res}^2}}/{{1/\sigma^{res}^2} + {1/\sigma^{kd}^2}}
%     \mu^{\mathit{ref}} &= w_1.(\mu^{\mathit{res}}+\mu^{\mathit{kd}}) +w_2.\mu^{\mathit{kd}},\\
%      \sigma^{\mathit{ref}} &= w_1{\sigma^{\mathit{res}}}w_1^T +  w_2{\sigma^{\mathit{kd}}}w_2^T + 2w_1{\sigma^{\mathit{res,kd}}}w_2^T, \\
%     % \mu^{\mathit{ref}} = 
%     % \dfrac{\mu^{\mathit{res}}{{\sigma^{\mathit{kd}}}^2}}{{\sigma^{\mathit{res}}}^2 + {\sigma^{\mathit{kd}}}^2} + \mu^{kd}, \\ %\hspace{3mm}  
%     % \ \ \ \sigma^{\mathit{ref}} = \dfrac{{{\sigma^{\mathit{res}}}^2}{{\sigma^{\mathit{kd}}}^2}}{{\sigma^{\mathit{res}}}^2 + {\sigma^{\mathit{kd}}}^2}.
%  \end{split} 
% \label{eq:IVW}
% \end{equation} 
% where $w_1,w_2$ are weight vectors defined in Equation \ref{}, $\sigma^{\mathit{kd}},\sigma^{\mathit{res,kd}}$ are covariance and cross-covariance matrices respectively. 
% We are given two trajectories $\mathbf{y}^{kd},\mathbf{y}^{kd}+\mathbf{y}^{res}$ with Gaussian distributions and want to merge them. We assume independent points on each trajectory and represent each point on the trajectories $\mathbf{y}^{kd},\mathbf{y}^{kd}+\mathbf{y}^{res}$ by $x,x'$ respectively.% as two element vectors of x and y coordinates. 
% Each point on $\mathbf{y}^{ref}$ is also shown by $z$. 
Hence, our goal is to find the weights $w,\widetilde{w}$ such that each point on the merged trajectory shown in Equation \ref{merge_equation} has the minimum variance. 
\begin{equation}\label{merge_equation}
     \mathbf{y}^{ref} = w\mathbf{y}^{kd}+\widetilde{w}(\mathbf{y}^{kd}+\mathbf{y}^{res}).
\end{equation}
Let's represent $\mathbf{y}^{kd}+\mathbf{y}^{res}$ by $\mathbf{y}^{ad}$ for brevity. The points on each trajectory are assumed temporally-independent. Weight matrices have the form of $w=diag(w_1,w_2),\widetilde{w}=diag(\widetilde{w}_1,\widetilde{w}_2)$ where $w_i,\widetilde{w}_i$ are scalar values. Note that $\mathbf{y}^{kd}$ is a prior for $\mathbf{y}^{res}$ and hence they are not independent.
The problem formulation is as follows:
\begin{align}
\begin{aligned}
         &\argmin_{w,\widetilde{w}} (\sigma^{ref}_{11},\sigma^{ref}_{22}),&\\
    &\text{ subject to: } \sigma_{ref} =  w\sigma^{kd}w^T + \widetilde{w}\sigma^{ad}\widetilde{w}^T + 2w\sigma^{{kd},{ad}}\widetilde{w}^T,\\
    & \hspace{16mm} w+\widetilde{w} = I_{2\times2},
\end{aligned}
\end{align}
where $\sigma^{kd},\sigma^{ad}$ are the covariance matrices and $\sigma^{{kd},{ad}}$ is the cross-covariance matrix. %By setting the derivatives of $\sigma_{z_{11}},\sigma_{z_{22}}$ with respect to $w,\widetilde{w}$ to zero,
Solving the constraint problem using Lagrangian multipliers leads to the following solution: 
\begin{align}
\begin{aligned}
w = \begin{bmatrix}
\dfrac{\sigma^{ad}_{11}-\sigma^{kd,ad}_{11}} {\sigma^{ad}_{11}+\sigma^{kd}_{11}-2\sigma^{kd,ad}_{11}} \vspace{2mm} \\
\dfrac{\sigma^{ad}_{22}-\sigma^{kd,ad}_{22}} {\sigma^{ad}_{22}+\sigma^{kd}_{22}-2\sigma^{kd,ad}_{22}}
\end{bmatrix},
\vspace{5mm}
\widetilde{w} = \begin{bmatrix}
\dfrac{\sigma^{kd}_{11}-\sigma^{kd,ad}_{11}} {\sigma^{ad}_{11}+\sigma^{kd}_{11}-2\sigma^{kd,ad}_{11}} \vspace{2mm} \\
\dfrac{\sigma^{kd}_{22}-\sigma^{kd,ad}_{22}} {\sigma^{ad}_{22}+\sigma^{kd}_{22}-2\sigma^{kd,ad}_{22}}
\end{bmatrix}.
\label{eq:ivw_weights}
\end{aligned}
\end{align}

\subsection{Multimodal Prediction and Loss Function}
Our approach can easily be extended to a multimodal predictor using Winner-Takes-All (WTA) loss (also known as oracle loss) approach \cite{wta}. To do so, the knowledge-aware model should generate multiple plausible future trajectories. RRB takes the KD predictions and finds the associated required residuals. Finally, the loss will be calculated, considering the closest mode to the ground truth.
Minimizing the log-likelihood will lead to the following loss function: 
\begin{equation}\label{loss}
    \begin{aligned}
    l(\theta) = -\sum_{n=1}^N \sum_{m=1}^M \mathbbm{1}(m=m^*)[
    \textrm{log} \,  p(\mathbf{x}^{p}_{n,m}|S_n,\mu^{\mathit{ref}}_{n,m}(\theta),\sigma^{\mathit{ref}}_{n,m}(\theta))],
    %\mathcal{N}
    \end{aligned}
\end{equation}
where $N$ and $M$ are number of samples and modes respectively, $\mathbbm{1}$ is the indicator function, and $m^*$ is the closest output to the ground truth in terms of $\ell_2$ distance.
Note that for the single modal case, the equation holds with $M=1$.

\subsection{Model Predictive Control} \label{sec:dmpc}
%@ismail
To add kinematic feasibility, in contrast with the previous works \cite{uber_kinematic,Conditional} which utilize a kinematic layer after computing control commands, we employ Model Predictive Control (MPC). This gives the model the flexibility to estimate the positions instead of finding the control commands which is beneficial in adding the domain-knowledge. MPC minimizes its cost function subject to a set of constraints. The state parameters for agent $i$ in time $t$ is $s_{t} = [x_{t},y_{t},\phi_{t},v_{t}]$ which consists of coordinates, orientation and speed. We denote control parameters acceleration and the steering angle by $ u_{t}=[a_{t},\gamma_{t}]$. The dynamics of the system is formulated using bicycle model $F_{bic}$ \cite{bicycle_model} which is shown to be sufficient for normal manoeuvres \cite{bicycle_enough}. Then, the MPC solves the following optimization problem:
\begin{align}
    \begin{aligned}
        \mathbf{y}^{p}_{t}= &\argmin_{s_{t:t+T_{p}},u_{t:t+T_{p}}} \sum_{j=1}^{T_{p}}  ||\ s_{t+j}[0:1] - \mathbf{y}_{t+j}^{ref}\ ||_2^2 + \lambda||\ u_{t+j}-u_{t+j-1}\ ||_2^2 \\
        &\text{ subject to: }  
         s_{t+1} = F_{bic}(s_{t}, u_{t}), \
        \ s_{0} = s_{init}, \ \ u_{min} < u_{t} < u_{max},
  \label{eq:mpc}
      \end{aligned}
\end{align}

 \iffalse
\begin{align}
    \begin{aligned}
        \mathbf{y}^{p}_{t}=&\argmin_{s_{t:t+T_{p}},u_{t:t+T_{p}}} \sum_{j=1}^{T_{p}}  g_{t+j}(s_{t+j},u_{t+j}); \\
        & g_{j}(s_{t},u_{t}) = ||\ s_{t}[0:1] - \mathbf{y}_{t}^{ref}\ ||_2^2 + \lambda||\ u_{t}-u_{t-1}\ ||_2^2 \\
        \text{ subject to:} \\ 
        & s_{t+1} = F_{bic}(s_{t}, u_{t}), \
        \ s_{0} = s_{init}, \ \ u_{min} < u_{t} < u_{max}
  \label{eq:mpc}
      \end{aligned}
\end{align}
\fi
where ${\bf y}^{ref}$ is the reference trajectory, $\lambda$ is a hyper-parameter and $u_{min}, u_{max}$ are minimum and maximum feasible control values respectively. %We desire to train the network end-to-end thus all blocks in the model should be differentiable. Since equation \ref{eq:mpc} is not differentiable, %(@ismail write why here?????), 
%the proposed solution in \cite{DeepMPC} is employed to solve it in a differentiable way.
\section{Experiments}
% baselines, the baselines I had for my presntaion in group-meeting: social-gan, social LSTM (maybe only one of them), kd + m, kd+res+m 
% size of data + prediction length
% ablation: res block: baysian, optimization
% base predictor: use NN and constraint it, next, train using one net, test with another
We evaluate our RRB against other baselines to test the following hypotheses: (1) RRB brings the advantages of both knowledge-driven and data-driven models thus, outperforms both types of models, (2) RRB structure merges knowledge-driven and data-driven models better than other fusion techniques, (3) adding MPC makes outputs more realistic by satisfying kinematic constraints, (4) our trained RRB can improve performance of different knowledge-driven models without the need for fine-tuning, %and (5) IVW-addition and residual confinement are essential parts of RRB for having more realistic outputs.
(5) IVW-addition and residual confinement are essential parts for more realistic outputs.
%(1) Does the proposed model bring the advantages of both knowledge-driven and data-driven worlds? We will investigate it by comparing the performance of our approach with both knowledge-driven and data-driven models. (2) Is the choice of RRB better than other possible alternatives? (3) How does the added MPC impact the results? (4) Is RRB robust with respect to the KD trajectory? In other words, can a trained RRB be applied to different knowledge-driven models? We will answer this question in the ablation studies.
\subsection{Dataset}
%\noindent {\bf Interaction dataset}.
We evaluate the proposed method on Interaction dataset \cite{interactiondataset}. It is a large-scale real-world dataset which consists of top-down scenes from intersections, highways, and roundabouts. 
The data is collected from three different continents (North America, Asia and Europe). It includes locations of dynamic agents such as vehicles and pedestrians for each frame of the 10 Hz downsampled video. The dataset also provides the static context of the scene.
Interaction dataset is challenging as it includes interactions between vehicles, different environments, and potentially multiple plausible predictions. 
We used the same settings as \cite{interactiondataset}. The observation and prediction lengths indicate the number of frames used to represent the past states and to be predicted, respectively. We have set observation lengths equal to $5$ frames (2.5 s) and and prediction length as $10$ frames (5 s). The dataset includes scenes with different numbers of samples. Hence, in order not to be overwhelmed by scenes with large number of samples, we report the average of performance of a model on the three categories regardless of the number of samples in each category.
Moreover, to study the model's generalization power, we consider two scenarios: the first scenario is scene-overfitting, in which all scenes exist in the training set, but 20\% of data is kept for the test set. In this scenario the model should be able to overfit on the scene and generalize on the interactions. We argue that to assess model generalization in terms of scene perception, another scenario named scene-generalization is required. In this scenario, the three following scenes, 'DR\_USA\_Intersection\_MA', 'DR\_USA\_Roundabout\_SR', 'DR\_CHN\_Merging\_ZS' are kept for the test set and the rest scenes are used in training. This challenging scenario reveals the performance of the models in a new environment. 
%To assess model's generalization power, we consider scene-generalization scenario where the three following scenes, 'DR\_USA\_Intersection\_MA', 'DR\_USA\_Roundabout\_SR', 'DR\_CHN\_Merging\_ZS' were kept for the test set and the rest scenes were used in training. This reveals the performance of the models in a new environment. 
\subsection{Implementation Details} \label{implementation-details}
The models are trained for 50 epochs with batch size of 32. We employed Adam optimizer \cite{adam} with the initial learning rate of 0.001, which is decreased by half every 10 epochs. The model is implemented using PyTorch \cite{pytorch}. Network's building blocks are MLP networks with the following hidden layers: History and interaction encoders have (32,32,64), KD encoder has (32,64) and the decoder has (256,128,128,64) hidden layers.

\subsection{Metrics and Baselines} \label{metrics-section}
%We used Average/final displacement error {\bf (ADE/FDE)} \cite{}, Road violation {\bf (RV)} and Cross track {\bf(CT)} metrics to evaluate models. 
The following metrics were used for the evaluation:
%\vspace{3}
\begin{enumerate}[leftmargin=*]
    \item Average/final displacement error {\bf (ADE/FDE)}. Average displacement error (ADE) and Final displacement error (FDE) are adopted as two common evaluation metrics. In the multimodal case, similar to the previous works \cite{social_gan,lee2017desire}, the closest mode to the ground truth is chosen. % among predicted modes.
    
    %Average/final displacement error {\bf (ADE/FDE)}. Average displacement error (ADE) is the average euclidean distance between the predicted points and the ground truth over all predicted time steps. % $ \sum_{i=1}^N \sum_{t=1}^{T_{p}} ||\mathbf{y}^p_t-\mathbf{x}_t^p||_2/(N*T_{p})$ where $N$ is the number of samples.
    %Final displacement error (FDE) is the  displacement error between the final predicted points at the end of the prediction horizon and the actual destinations. %: $ \sum_{i=1}^N ||\mathbf{y}^p_{T_{p}}-\mathbf{x}_{T_{p}}^p||_2/N$.
    
    \item Road violation {\bf (RV)}. Inspired by \cite{uber-scene}, we define this metric as the percentage of average number of points predicted in the off-road area. For the multimodal case, we average RVs for the modes weighted by their probability. RV measures the feasibility of predictions concerning the scene.
    
    \item  Cross track {\bf(CT)}. Cross track is the distance between the actual destination and the final point on the retimed predicted sequence by the ground truth speed profile \cite{gong2004methodology}. In other words, cross track is the distance between the actual destination and the final point on the retimed predicted sequence.
    Cross track metric is able to express the spatial effectiveness of the model, excluding temporal aspects.  
    %To calculate cross track error, predicted sequence is retimed by ground truth speed profile. Cross track is the distance between actual destination and the final point on the retimed predicted sequence. Cross track metric is able to express the spatial effectiveness of the model excluding temporal aspects.  
    %While ADE and FDE indicate how well the model performs with respect to the ground truth, they consist of both temporal and spatial features. Cross track metric \cite{gong2004methodology} is able to express the spatial effectiveness of the model excluding temporal aspects.  
\end{enumerate}
% \noindent  {\bf Scene loss (SL)}. Scene loss aims at finding out how well the model performs on the scene part. Inspired by \cite{uber-scene}, we define this metric as the average number of points predicted in the off-road area.
To demonstrate the effectiveness of RRB, we compare the results with the following baselines: 
%\begin{itemize}

\noindent \textit{Naive baselines}: We used Kalman filter as a linear prediction {\bf (Lin)}. We also report Constant velocity {\bf(CV)} \cite{the-simpler-the-better} as another naive baseline.

\noindent \textit{Knowledge-driven baselines}:  In order to form the KD trajectory explained in section \ref{knowledge-aware-section}, we utilize different strategies to predict vehicle velocity. We name Constant velocity strategy \cite{the-simpler-the-better} as {\bf KD1}, and Leader follower \cite{leader-follower} which tackles interaction between agents as {\bf KD2}. 
    
\noindent \textit{Data-driven baselines:} For the data-driven models that address agent-agent interactions, we report Social LSTM {\bf (S-LSTM)} \cite{social_lstm} , Social Attention {\bf (S-ATT)} \cite{social-att} (numbers reported from \cite{social-wagdat}), 
and Social GAN {\bf (S-GAN\_M)} \cite{social_gan}. We report Social WaGDAT {\bf (S-WaGDAT\_M)} \cite{social-wagdat} (numbers reported from their paper) as a state-of-the-art work that models both scene and interactions. We also report the performance of our encoder-decoder neural network {\bf(EDN)} similar to Figure \ref{fig:residual_estimator_model} trained with equation \ref{loss} as the loss function. We replaced the knowledge-driven model in Figure \ref{fig:residual_estimator_model} with a convolutional neural network to represent a fully data-driven model. Note that the multimodal baselines' names end with '\_M'.
    
\noindent \textit{Mixed baselines:} We implemented the road loss {\bf(RL)} \cite{uber-scene} to impose the scene constraint to the model. It adds the road loss to the NN model to help it learn the drivable and non-drivable regions. 
%As an alternative to the RRB model, one can use Bayes rule {\bf(Bayesian)}. Consider the KD1 output as the prior and NN's output as the likelihood. Hence, the posterior can be directly achieved by Bayes rule according to the following equation:
    % \begin{equation}
    %     p(y^{\mathit{p}}_{i,t}|x^p_{i,t},S_t) = p_{\scriptscriptstyle{\mathit{NN}}}(x^p_{i,t}|y^{\mathit{p}}_{i,t},S_t)*p_{\scriptscriptstyle{ \mathit{KD}}}(y^{p}_{i,t}|S_t)
    % \end{equation}
% As an alternative to the RRB model, one can use a Variance-based Integration {\bf(VI)} between the KD and data-driven predictions by employing IVW method \cite{ivw}. For each sample: 
% \begin{equation}
%     \mu^{p} = \dfrac{\mu^{\scriptscriptstyle{\mathit{EDN}}} {{\sigma^{\scriptscriptstyle{\mathit{KD}}}}^2} + \mu^{\scriptscriptstyle{\mathit{KD}}} {{\sigma^{\scriptscriptstyle{\mathit{EDN}}} }^2}}{{\sigma^{\scriptscriptstyle{\mathit{EDN}}}}^2 + {\sigma^{\scriptscriptstyle{\mathit{KD}}}}^2} 
% \end{equation}
% where $\mu$ and $\sigma$ are mean and variance respectively. Note that this is equivalent to using Bayes rule while KD output is taken as the prior and EDN as the likelihood.
We also consider two alternatives to the RRB model. The first one is to use a Variance-based Integration {\bf(VI1)} between the KD1 and data-driven prediction EDN by employing IVW method \cite{ivw}. Note that this is equivalent to using Bayes rule while KD output is taken as the prior and EDN as the likelihood. The second approach is to replace the KD variance in VI1 by a fixed hyperparameter {\bf(VI2)}. 

Our proposed solution is RRB model {\bf (RRB)} explained in section \ref{rrb-section} which uses KD1 method. We express the effectiveness of our method on other KD predictions in the ablation study. We also demonstrate the performance of our model in the multimodal case {\bf(RRB\_M)} with 2 modes and when constrained by the MPC {\bf (RRB\_M+MPC)}. Moreover, in the ablation study, we report the performance of RRB while IVW-addition is replaced by simple addition {\bf(A-RRB)} and also a non-confined RRB model {\bf (NC-RRB)}.

\subsection{Results}

\begin{table}[t]%[width=0.1\textwidth]
\caption{Quantitative results of baselines in scene-generalization scenario. ADE/FDE and CT are reported in meters. The lower the better for all metrics.}
\centering
    \begin{tabular}{c  c | lll}
    \toprule
    &\multirow{2}{10em}{Models}  & \multicolumn{3}{c}{Scene-generalization} \\ \cmidrule(r){3-5} \cmidrule(r){3-5}
     &&  \multicolumn{1}{c}{ADE/FDE} & \multicolumn{1}{c}{RV} & \multicolumn{1}{c}{CT} %& \multicolumn{1}{c}{ADE/FDE} & \multicolumn{1}{c}{SL} & \multicolumn{1}{c}{CT} 
     \\ \midrule
    \multirow{2}{5em}{Naive} \
    &Lin & 4.13 / 8.77 & 34 & 3.6 \\ 
    &CV \cite{the-simpler-the-better} & 3.12 / 7.34 & 24 & 3.19 \\ 
      \midrule
    \multirow{2}{5em}{Knowledge-driven} \
     %\cline{2-8}
    %&CV \cite{the-simpler-the-better} &2.78 / 6.57& 0.3 &5.14 & 3.10 / 7.31&0.49 & 5.24 \\ \cline{2-8}
    &KD1 \cite{the-simpler-the-better,benz} & 2.92 / 6.62 & {\bf0} & 1.87 \\ %\cline{2-8}
    &KD2 \cite{leader-follower,benz}   & 2.85 / 6.55 & {\bf0} & {\bf 1.74} \\ %\cline{2-8}
    \midrule
    %Social-LSTM  & 2.17 & 5.12 &  & 2.69 & 4.64 &  & 2.19 & 3.87 &  \\ \hline
    \multirow{3}{5em}{Data-driven}
    &S-LSTM \cite{social_lstm}  &  2.85 / 7.17 &  72
     & 4.26 \\ %\cline{2-8}
    &S-GAN\_M \cite{social_gan}  & 2.34 / 5.82 & 52 & 5.71 \\ %\cline{2-8}
    &EDN & 2.78 / 6.7 & 7 & 3.4 \\ %\cline{2-8}
    %&EDN\_M &   1.85 / 4.56 & 7 & 2.64 \\ 
    \midrule
    \multirow{5}{5em}{Mixed approaches}
    % &Baysian &2.31 / 5.51 &0.01 &3.66 & 2.79 / 6.48 & 0.01 &4.22 \\
    &VI1 & 2.52 / 6.27  & 3 & 2.9  \\%\cline{2-8}
    &VI2 & 2.62 / 6.38 & 2 & 2.32 \\
    & RL \cite{uber-scene} & 2.56 / 6.40 &  9 & 3.15  \\ %\cline{2-8}
    %  &VI &2.63 / 6.21 & 0.01 &4.49&2.58 / 6.27 & 0.04 & 4.24\\ %\cline{2-8}
     &RRB& 2.44 / 6.04 & {\bf 0} & {1.98} \\
    %&our RRB &  /   &    & 3.85 & 2.65 / 6.44 & 0.02 & {4.65} \\
    %\cline{2-8}
    &RRB\_M &  {2.15} / 5.08 & {\bf 0} & 1.81 \\ %\cline{2-8}
    &RRB\_M+MPC & {\bf2.13} / {\bf 5.02} & {\bf 0} & 1.81 \\
    \bottomrule
    \end{tabular}
    \label{scene-generalization-results}
\end{table}

\begin{table}[t]
    \caption{Quantitative results of baselines in scene-overfitting scenario. ADE/FDE are in meters.}
    
        \centering
        \begin{tabular}{c | c}
        \toprule
        Models & ADE/FDE \\
        \midrule
        CV \cite{the-simpler-the-better} & 2.80 / 6.59 \\
        KD1 \cite{the-simpler-the-better,benz} &2.59 / 6.00 \\
        KD2 \cite{leader-follower,benz} &2.53 / 5.83 \\
        S-LSTM \cite{social_lstm} & 2.33 / 4.52 \\ %1.89 / 4.70 \\
        EDN & 1.81 / 4.05 \\
        S-ATT \cite{social-att} & 2.29 / 4.25 \\
        S-GAN\_M \cite{social_gan} & 2.12 / 4.20 \\
        S-WaGDAT\_M \cite{social-wagdat} & 1.62 / {\bf 3.35} \\
        %RL \cite{uber-scene} & 1.83 / 4.04 \\
        %EDN\_M & {\bf 1.21} / {\bf 2.59} \\
        RRB\_M & {\bf 1.49} / 3.68 \\
        \bottomrule
        \end{tabular}
        \label{scene-overfitting-results}
    \end{table}
Table \ref{scene-generalization-results} provides the results in scene-generalization scenario. Knowledge-driven models outperform other models in terms of RV and CT metrics since they can perceive scene well and also generalize to new environment. The data-driven approaches have better ADE/FDE as they can learn interactions from data. Although EDN performs well in scene-overfiting scenario (shown in Table \ref{scene-overfitting-results}), it fails in generalizing to the new environment leading to high RV and CT values. 
The large gap between the performance of data-driven methods in scene-overfitting and the scene-generalization scenarios reveals the limited generalization power of solely data-driven approaches, which can be mitigated by leveraging domain-knowledge.
The proposed RRB model outperforms knowledge-driven models as it learns vehicle-vehicle interactions from data and goes beyond average behavior. Compared to the data-driven models, RRB can better generalize to new environments due to utilization of the scene knowledge. Compared with the mixed approaches,% RRB has much less off-road predictions because of the direct use of scene knowledge. 
RRB outperforms both VI1 and VI2 %in both scenarios 
because as opposed to them that merge two independent models, RRB generates residuals conditioned on KD output. Our method has zero RV because of the realistic residual scale which prevents off-road predictions. Note that RL could not improve EDM performance as it is very sensitive to the choice of hyperparameters and has many local minima.

The possibility of having multiple outputs is investigated by increasing the number of modes of RRB model to 2. The results approve that RRB\_M is able to successfully capture multiple modes while it provides the required residuals for each mode individually leading to a multimodal knowledge-aware model. 
MPC is employed in RRB\_M+MPC in order to ensure the kinematic-feasibility of the outputs. The results show that kinematic constraints are satisfied without loss of performance. Hence, the model better approximates human behavior.

The results of previous works in scene-overfitting scenario are reported in Table \ref{scene-overfitting-results}. Our RRB\_M model outperforms previous works and especially the recent S-WaGDAT\_M \cite{social-wagdat} model %The RRB\_M has competitive results with other models 
which expresses the ability of the model to learn from data.
Note that RRB\_M could successfully improve performance of KD1 by adding the residuals. %Note that the cost of having realistic outputs in RRB\_M is that the accuracy in scene-overfitting scenario is lower-bounded by EDN\_M because of confining the residuals. 

%our EDN model has competitive performance with other baselines. Its low CT reflects its ability to perceive the scene and ADE/FDE numbers show the effectiveness of the agent-agent interaction module.

% In the mixed approaches, RL could not improve EDM performance. It is very sensitive to the choice of hyperparameters and has many local minima.
% VI utilizes data and scene knowledge. The data helped the model to improve ADE/FDE concerning K1, and the knowledge reduced EDN's number of off-road samples.

%****performance vs data size and prediction length*****

\begin{figure*}[t]
\centering
\subfloat{\includegraphics[width=1.36in]{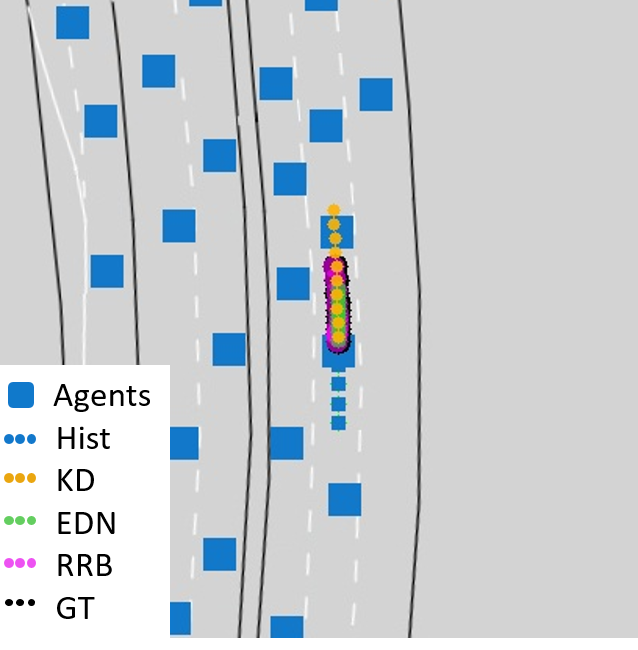}
\label{fig_r5}}
\subfloat{\includegraphics[width=1.35in]{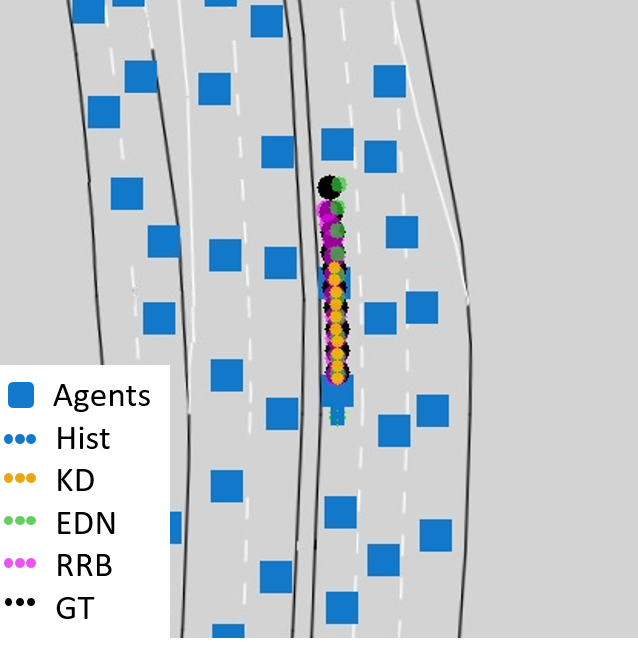}
\label{fig_r6}}
\subfloat{\includegraphics[width=1.35in]{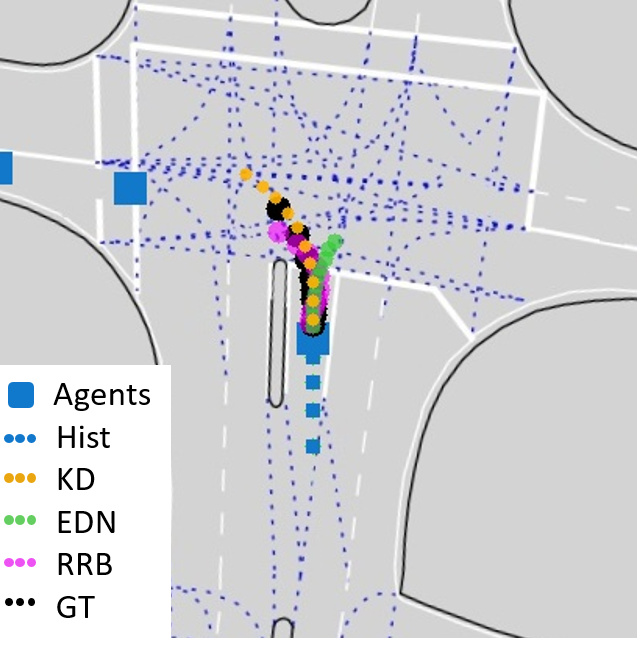}
\label{fig_r7}}
\subfloat{\includegraphics[width=1.35in]{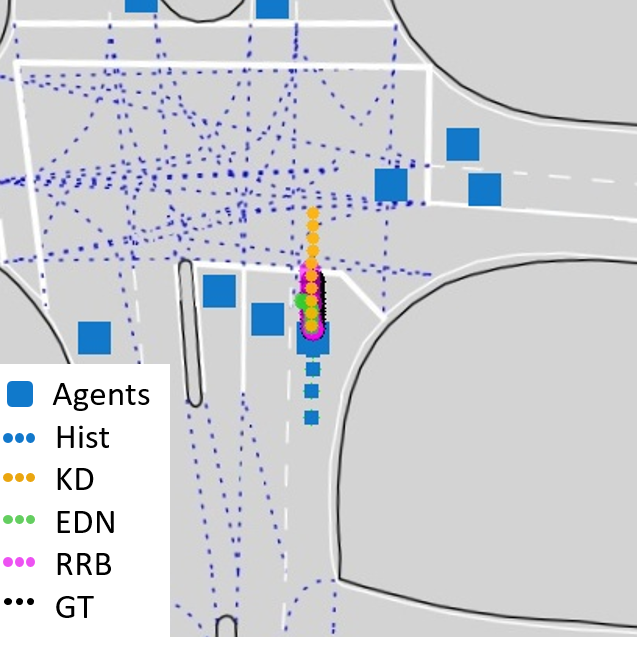}
\label{fig_r8}}\\
\subfloat{\includegraphics[width=1.35in]{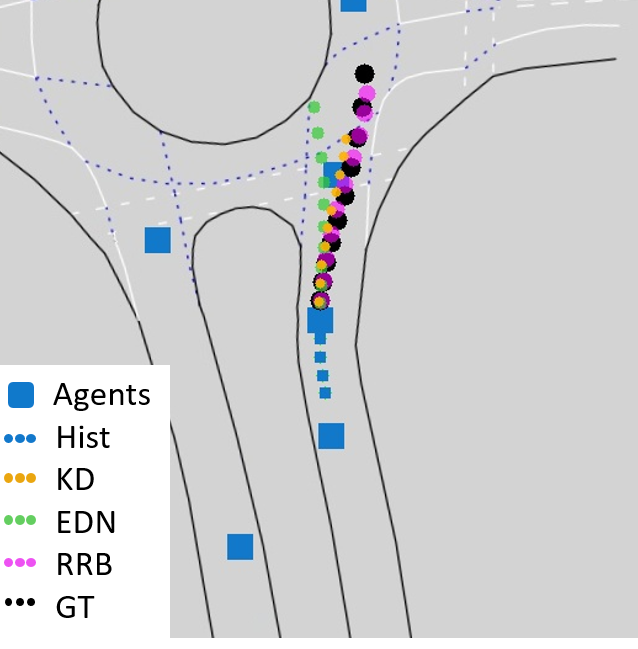}
\label{fig_r1}}
\subfloat{\includegraphics[width=1.36in]{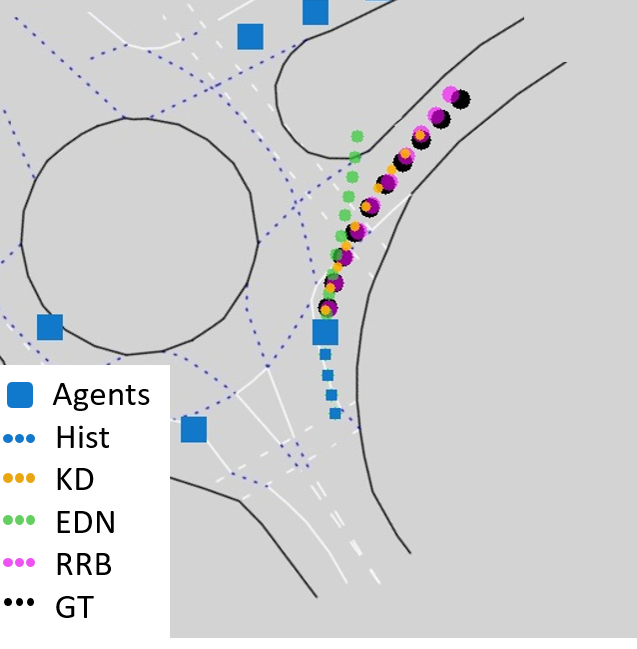}
\label{fig_r2}}
\subfloat{\includegraphics[width=1.35in]{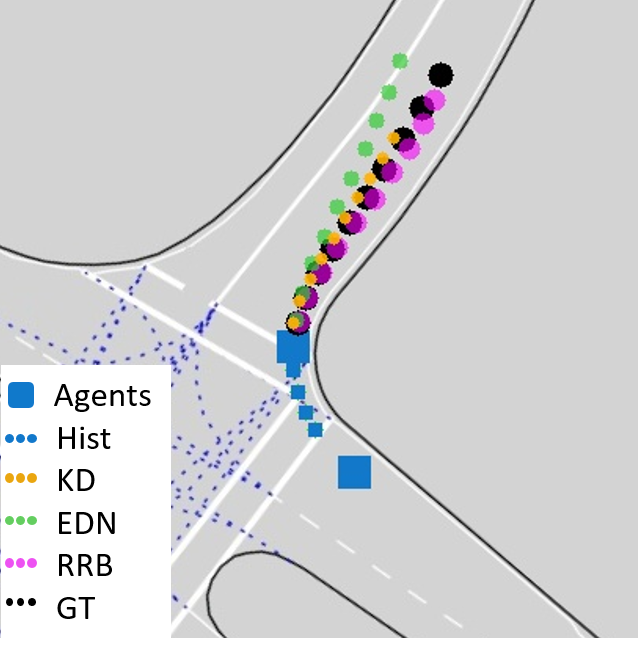}
\label{fig_r3}}
\subfloat{\includegraphics[width=1.35in]{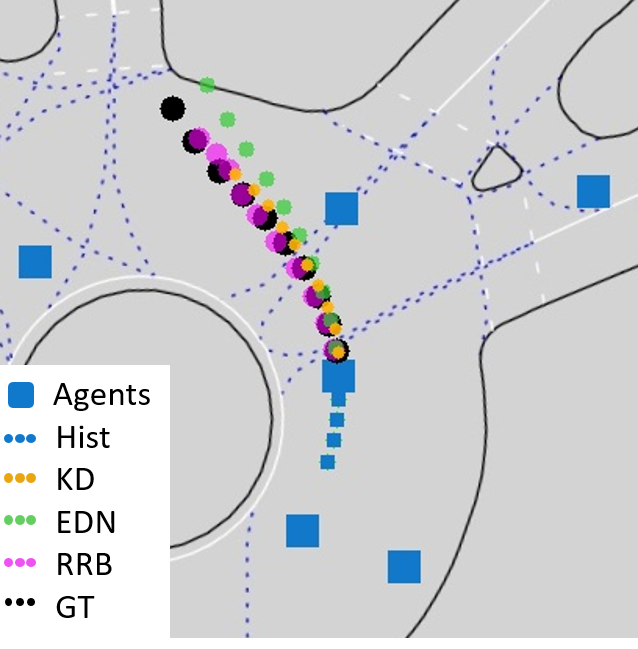}
\label{fig_r4}} 

%\subfloat{\includegraphics[width=1.33in]{figures/ablation3.png}%
%\label{fig_r3}}
% \hfil
%\caption{Images of each of the four roundabouts in the dataset: Avenue de la Gotta \ref{fig_r1}, Route de la Gare \ref{fig_r2}, Route de la Pierre \ref{fig_r3}, and EPFL Route Cantonale \ref{fig_r4}. }
\caption{Qualitative results of different baselines. Given the scene and history (Hist), the models predict the future positions. The ground truth (GT) is shown in black. The encoder-decoder neural network (EDN) captures the interactions while is prone to create unrealistic outputs shown in first row. Knowledge-driven (KD) model has realistic predictions as it uses the scene knowledge but cannot reason about the interactions among vehicles as shown in the second row. Our proposed RRB predicts realistic and interaction-aware outputs.}
\label{fig:qualitative}
\end{figure*}

We visualize the outputs of different models in Figure \ref{fig:qualitative}. 
The first row images visualize the cases where KD prediction is not accurate because of the missing interactions. RRB complements KD trajectory by accounting for interactions in the residuals. The second row images emphasize on realisticity of the predictions. 
Although EDN can reason about interactions, it can have unrealistic predictions with respect to the scene especially in a new environment. However, the realistic elements of RRB preserve the feasibility of the final output as shown in all images.
%The figures show the feasibility and accuracy of RRB predictions.  while

\subsection{Ablation Study}\label{ablation-section}
% ablation: res block: baysian, optimization
% base predictor: use NN and constraint it, next, train using one net, test with another

In this section, a set of ablation studies are performed to shed light on the effectiveness of different parts of the model. First, we want to study the robustness of RRB concerning the different knowledge-driven models. We report two knowledge-driven methods in Table \ref{scene-generalization-results} and report the RRB performance while KD1 was employed. Without retraining the model, we replaced KD1 with Lin (Lin+RRB) and KD2 (KD2+RRB) models. The results are expressed in Table \ref{ablation_kd}, where the performance of the knowledge-driven model is reminded in the parenthesis. It shows that RRB could successfully improve the performance of the models in all metrics. This experiment shows the robustness and effectiveness of the proposed RRB for different KD trajectories.
%impact of using different knowledge-driven predictions. Two robust and non-robust RRB models are trained. The robust model {(RRB\_R)} used the noise-injection approach explained in Section \ref{knowledge-aware-section} while the non-robust model {(RRB\_NR)} didn't employ that. Without retraining the models, we replace our KD model with Kalman filter {(KF)} and leader following {(LF)} models as two other knowledge-based approaches. The results are expressed in table \ref{ablation_kd}. It shows RRB\_NR is not generalizable to new KD models. However, RRB\_R is much more robust to the choice of KD model. This experiment approves the robustness and effectiveness of the proposed RRB for different KD trajectories.

%that replacing the Kalman filter by our neural network predictor (NN+KD+Res) slightly affects the performance, which means that our approach is effective for different knowledge-aware models. 

%In order to investigate generalizability of RRB, we replaced 
%we assessed the perfomance of our model while NN is used to create the KD trajectory but RRB is trained with kalman filter as base predictor and tested the model while NN was used as the base predictor. As can be seen, the performance didn't change noticeably which shows after the model is trained, it can be used with different base networks.
The second ablation study aims at assessing realistic elements of RRB which are the realistic parameter ${\bf c}$ and IVW-addition. We trained the model without the limitation on the range of RRB as non-confined RRB (NC-RRB). In addition, IVW-addition can be replaced with simply adding the residuals to the KD prediction which is reported as A-RRB. This is equivalent to setting $w=diag(0,0),\widetilde{w}=diag(1,1)$ in Equation \ref{merge_equation}. 
The results are expressed in Table \ref{ablation_rrb}. 
Non-confined model has more freedom in improving the KD trajectory hence achieves a better ADE/FDE. But it is prone to unrealistic predictions with large RV error. Also, neglecting uncertainties in A-RRB will increase RV and CT as the network utilized uncertain residuals. These experiments approve that the added elements play important roles in having realistic predictions.

\begin{table}[t]
\caption{Results of the two ablation studies on robustness and realistic elements of the proposed RRB}
    \begin{subtable}[h]{0.45\textwidth}
    \caption{Assessing the robustness of RRB with respect to different knowledge-driven models. RRB is not trained in this experiment but we used the fixed model in Table \ref{scene-generalization-results}.}
        \centering
        \begin{tabular}{c lll}
        \toprule
        \multirow{1}{4em}{Models}  & \multicolumn{3}{c}{Scene-generalization} \\ \cmidrule(r){2-4}
         & \multicolumn{1}{c}{ADE/FDE} & \multicolumn{1}{c}{RV} & \multicolumn{1}{c}{CT}  \\ \midrule
         Lin & 4.13 / 8.77 & 34  & 3.6 \\ %\midrule
        Lin+RRB  & 3.11 / 7.50 & 2 & 3.59  \\
        KD2  & 2.85 / 6.55 &  {0}  &  1.74 \\               
        KD2+RRB  & 2.49 / 6.14  & 0 & 2.04 \\
                                   
        \bottomrule
        \end{tabular}
        \label{ablation_kd}
    \end{subtable}
    \hfill
    \begin{subtable}[h]{0.45\textwidth}
    \caption{Ablation study on the two realistic elements of the proposed RRB, residual confinement and IVW-addition block.}
        \begin{tabular}{c lll}
        \toprule
        \multirow{1}{4em}{Models}  & \multicolumn{3}{c}{Scene-generalization} \\ \cmidrule(r){2-4} 
         & \multicolumn{1}{c}{ADE/FDE} & \multicolumn{1}{c}{RV} & \multicolumn{1}{c}{CT}  \\ \midrule
        NC-RRB & 2.28 / 5.59 & 10 & 1.94 \\ %\hline
        A-RRB  & 2.47 / 6.08 & 3 & 2.07 \\ %\hline
        RRB & 2.44 / 6.04 & 0 & 1.98  \\
        \bottomrule
        \end{tabular}
        \label{ablation_rrb}
    \end{subtable}
\label{ablation}

\end{table}

%\subsection{Kinematic feasibility}
%using an MPC after training the network {\bf (KD+R+MPC)}, and finally our proposed method {\bf (KD+R+DMPC)}.
% \section{Discussion}
% In this paper we argue that neither solely knowledge-driven nor purely data-driven models can provide realistic and accurate vehicle trajectory predictions.
% Our experiments demonstrates that knowledge-driven models have inferior performance but can generalize to new scenes. On the other hand, although data-driven models provide more accurate predictions, they tend to have unrealistic predictions in new scenes. Hence, our work promotes adding generalizability assessment to the common evaluation methods for data-driven models, more specifically we evaluated feasibility of the predictions in new scenes. Our proposed RRB borrows the best of both words and provides both precise and generalizable predictions. 

% Our experiments show the weaknesses of solely knowledge-driven or data-driven works. (as is a discussion, maybe have to consider experiments first)
% - Results are realistic, but previous works suffer from that, esp NN
% - Find a middle solution, Have the benefits of both
% - our work promotes knowledge-drivens to improve their accuracy and encourages data-driven guys to consider safety and generalization.

\section{Conclusions and Future Work}
In this paper, we addressed the safety-critical task of vehicle trajectory prediction (also known as microscopic traffic modeling). 
We argue that neither solely knowledge-driven nor purely data-driven models can provide generalizable and accurate predictions.
Given a set of experiments on a real-world dataset, our experiments demonstrate that knowledge-driven models have inferior performance but can generalize to new scenes. On the other hand, although data-driven models provide more accurate predictions, they tend to have unrealistic predictions in new scenes. 
%Hence, our work promotes adding generalizability assessment to the common evaluation methods for data-driven models. 
%In this work, we evaluated feasibility of the predictions in new scenes in addition to the accurateness.
Our proposed solution, named RRB, effectively merges knowledge-driven with data-driven models by finding residuals required to be added to the knowledge-driven prediction in order to model human stochastic behavior. It leverages physically-constrained Inverse-variance weighting approach to build realistic and confident residuals. To further make the predictions realistic, we employed a Model Predictive Control (MPC) to bring kinematic constraints to the final output.
RRB outperforms all other counterparts in accuracy and generalizability. 
As future work, we will study the effectiveness of RRB when employing more complex knowledge-driven methods on a variety of agents, including pedestrians. Pedestrians are not constrained to the roads but respect specific social rules.  We can take the existing knowledge-driven models such as “Social force” \cite{social-forces} as prior predictions and learn residuals required to better capture pedestrians dynamics. Moreover, our approach can help in complex multi-agent environments where there exist vehicles, pedestrians and cyclists each one with its own constraints and dynamics. In such scenarios, RRB can benefit from using strong knowledge-driven priors for each category and provide safe and accurate data-driven residuals. We hope that our work will pave the way to more methods combining the best of knowledge and data driven approaches.

\section{Acknowledgements}
This project was funded by Honda R{\&}D Co., Ltd and the European union's Horizon 2020 research and innovation program under the Marie Skłodowska-Curie grant agreement N 754354. Also, we would like to thank the reviewers for their helpful comments.
% The proposed approach can be extended to applications other than trajectory prediction that potentially have knowledge-based and data-driven solutions like image classification and pose estimation. 

% \section{Broader impact}
% Autonomous vehicles, thanks to AI, already share the road with human-driven vehicles, which is a critical task for safety. State-of-the-art trajectory prediction works only focus on data-based approaches, - our work raised the questions of the feasibility of the output of these models, but also their ability to generalize. One approach that we implement is to restrict the prediction area for the Neural Network, constraining it’s output to realistic and feasible trajectories .Extensive experiences on real and large-scale data contains naturalistic movements of various traffic participants in a variety of highly interactive driving scenarios from different countries strongly indicated the advantages of our method.

\clearpage

\bibliographystyle{splncs04}
\bibliography{egbib}

\begin{thebibliography}{10}
\providecommand{\url}[1]{\texttt{#1}}
\providecommand{\urlprefix}{URL }
\providecommand{\doi}[1]{https://doi.org/#1}

\bibitem{social_lstm}
Alahi, A., Goel, K., Ramanathan, V., Robicquet, A., Fei-Fei, L., Savarese, S.:
  Social lstm: Human trajectory prediction in crowded spaces. In: The IEEE
  Conference on Computer Vision and Pattern Recognition (CVPR) (June 2016)

\bibitem{alahi2017learning}
Alahi, A., Ramanathan, V., Goel, K., Robicquet, A., Sadeghian, A.A., Fei-Fei,
  L., Savarese, S.: Learning to predict human behavior in crowded scenes. In:
  Group and Crowd Behavior for Computer Vision, pp. 183--207. Elsevier (2017)

\bibitem{an2015space}
An, L., Tsou, M.H., Crook, S.E., Chun, Y., Spitzberg, B., Gawron, J.M., Gupta,
  D.K.: Space--time analysis: Concepts, quantitative methods, and future
  directions. Annals of the Association of American Geographers
  \textbf{105}(5),  891--914 (2015)

\bibitem{cooperation-aware}
Bae, S., Saxena, D., Nakhaei, A., Choi, C., Fujimura, K., Moura, S.J.:
  Cooperation-aware lane change maneuver in dense traffic based on model
  predictive control with recurrent neural network. arXiv
  \textbf{abs/1909.05665} (2019)

\bibitem{waymo}
{Bansal}, M., {Krizhevsky}, A., {Ogale}, A.S.: Chauffeurnet: Learning to drive
  by imitating the best and synthesizing the worst. In: Robotics: Science and
  Systems XV. vol.~15 (2019)

\bibitem{borghesi2020improving}
Borghesi, A., Baldo, F., Milano, M.: Improving deep learning models via
  constraint-based domain knowledge: a brief survey. arXiv preprint
  arXiv:2005.10691  (2020)

\bibitem{human-centered}
Chen, Y., Hu, C., Wang, J.: Human-centered trajectory tracking control for
  autonomous vehicles with driver cut-in behavior prediction. IEEE Transactions
  on Vehicular Technology  \textbf{68}(9),  8461--8471 (2019)

\bibitem{ivw}
Cochran, W.G.: The combination of estimates from different experiments.
  Biometrics  \textbf{10}(1),  101--129 (1954)

\bibitem{coscia2018long}
Coscia, P., Castaldo, F., Palmieri, F.A., Alahi, A., Savarese, S., Ballan, L.:
  Long-term path prediction in urban scenarios using circular distributions.
  Image and Vision Computing  \textbf{69},  81--91 (2018)

\bibitem{coscia2016point}
Coscia, P., Castaldo, F., Palmieri, F.A., Ballan, L., Alahi, A., Savarese, S.:
  Point-based path prediction from polar histograms. In: 2016 19th
  International Conference on Information Fusion (FUSION). pp. 1961--1967. IEEE
  (2016)

\bibitem{hondakf}
Cosgun, A., Ma, L., Chiu, J., Huang, J., Demir, M., Anon, A.M., Lian, T.,
  Tafish, H., Al-Stouhi, S.: Towards full automated drive in urban
  environments: A demonstration in gomentum station, california. In: 2017 IEEE
  Intelligent Vehicles Symposium (IV). pp. 1811--1818. IEEE (2017)

\bibitem{uber_kinematic}
Cui, H., Nguyen, T., Chou, F.C., Lin, T.H., Schneider, J., Bradley, D., Djuric,
  N.: Deep kinematic models for physically realistic prediction of vehicle
  trajectories. arXiv preprint arXiv:1908.00219  (2019)

\bibitem{uber_multimodal}
Cui, H., Radosavljevic, V., Chou, F.C., Lin, T.H., Nguyen, T., Huang, T.K.,
  Schneider, J., Djuric, N.: Multimodal trajectory predictions for autonomous
  driving using deep convolutional networks. In: 2019 International Conference
  on Robotics and Automation (ICRA). pp. 2090--2096. IEEE (2019)

\bibitem{convolution-pool}
Deo, N., Trivedi, M.M.: Convolutional social pooling for vehicle trajectory
  prediction. 2018 IEEE/CVF Conference on Computer Vision and Pattern
  Recognition Workshops (CVPRW) pp. 1549--15498 (2018)

\bibitem{fang2017object}
Fang, Y., Kuan, K., Lin, J., Tan, C., Chandrasekhar, V.: Object detection meets
  knowledge graphs  (2017)

\bibitem{francca2014fast}
Fran{\c{c}}a, M.V., Zaverucha, G., Garcez, A.S.d.: Fast relational learning
  using bottom clause propositionalization with artificial neural networks.
  Machine learning  \textbf{94}(1),  81--104 (2014)

\bibitem{posterior}
Ganchev, K., Gillenwater, J., Taskar, B., et~al.: Posterior regularization for
  structured latent variable models. Journal of Machine Learning Research
  \textbf{11}(Jul),  2001--2049 (2010)

\bibitem{garcez2019neural}
Garcez, A.d., Gori, M., Lamb, L.C., Serafini, L., Spranger, M., Tran, S.N.:
  Neural-symbolic computing: An effective methodology for principled
  integration of machine learning and reasoning. arXiv preprint
  arXiv:1905.06088  (2019)

\bibitem{gong2004methodology}
Gong, C., McNally, D.: A methodology for automated trajectory prediction
  analysis. In: AIAA Guidance, Navigation, and Control Conference and Exhibit.
  p.~4788

\bibitem{social_gan}
Gupta, A., Johnson, J., Fei-Fei, L., Savarese, S., Alahi, A.: Social gan:
  Socially acceptable trajectories with generative adversarial networks. In:
  Proceedings of the IEEE Conference on Computer Vision and Pattern
  Recognition. pp. 2255--2264 (2018)

\bibitem{hartung2011statistical}
Hartung, J., Knapp, G., Sinha, B.K.: Statistical meta-analysis with
  applications, vol.~738. John Wiley \& Sons (2011)

\bibitem{social-forces}
Helbing, D., Molnar, P.: Social force model for pedestrian dynamics. Physical
  Review E  \textbf{51} (05 1998). \doi{10.1103/PhysRevE.51.4282}

\bibitem{surround-vehicle}
{Jeong}, Y., {Kim}, S., {Yi}, K.: Surround vehicle motion prediction using
  lstm-rnn for motion planning of autonomous vehicles at multi-lane turn
  intersections. IEEE Open Journal of Intelligent Transportation Systems
  \textbf{1},  2--14 (2020). \doi{10.1109/OJITS.2020.2965969}

\bibitem{kalman}
Kalman, R.E.: A new approach to linear filtering and prediction problems
  (1960)

\bibitem{kani2017dr}
Kani, J.N., Elsheikh, A.H.: Dr-rnn: A deep residual recurrent neural network
  for model reduction. arXiv preprint arXiv:1709.00939  (2017)

\bibitem{karpatne2017physics}
Karpatne, A., Watkins, W., Read, J., Kumar, V.: Physics-guided neural networks
  (pgnn): An application in lake temperature modeling. arXiv preprint
  arXiv:1710.11431  (2017)

\bibitem{learning-based}
Kazemi, H., Mahjoub, H.N., Tahmasbi-Sarvestani, A., Fallah, Y.P.: A
  learning-based stochastic mpc design for cooperative adaptive cruise control
  to handle interfering vehicles. IEEE Transactions on Intelligent Vehicles
  \textbf{3}(3),  266--275 (2018)

\bibitem{keyvan2016categorization}
Keyvan-Ekbatani, M., Knoop, V.L., Daamen, W.: Categorization of the lane change
  decision process on freeways. Transportation research part C: emerging
  technologies  \textbf{69},  515--526 (2016)

\bibitem{adam}
Kingma, D.P., Ba, J.: Adam: A method for stochastic optimization.
  arXiv:1412.6980  (2014)

\bibitem{bicycle_model}
Kong, J., Pfeiffer, M., Schildbach, G., Borrelli, F.: Kinematic and dynamic
  vehicle models for autonomous driving control design. In: 2015 IEEE
  Intelligent Vehicles Symposium (IV). pp. 1094--1099. IEEE (2015)

\bibitem{bicycle_enough}
Kong, J., Pfeiffer, M., Schildbach, G., Borrelli, F.: Kinematic and dynamic
  vehicle models for autonomous driving control design. In: 2015 IEEE
  Intelligent Vehicles Symposium (IV). pp. 1094--1099. IEEE (2015)

\bibitem{kothari2020human}
Kothari, P., Kreiss, S., Alahi, A.: Human trajectory forecasting in crowds: A
  deep learning perspective. arXiv preprint arXiv:2007.03639  (2020)

\bibitem{lake2015human}
Lake, B.M., Salakhutdinov, R., Tenenbaum, J.B.: Human-level concept learning
  through probabilistic program induction. Science  \textbf{350}(6266),
  1332--1338 (2015)

\bibitem{lake2017building}
Lake, B.M., Ullman, T.D., Tenenbaum, J.B., Gershman, S.J.: Building machines
  that learn and think like people. Behavioral and brain sciences  \textbf{40}
  (2017)

\bibitem{cnn-lecun}
LeCun, Y., Bottou, L., Bengio, Y., Haffner, P.: Gradient-based learning applied
  to document recognition. Proceedings of the IEEE  \textbf{86}(11),
  2278--2324 (1998)

\bibitem{desire}
Lee, N., Choi, W., Vernaza, P., Choy, C., H.~S.~Torr, P., Chandraker, M.:
  Desire: Distant future prediction in dynamic scenes with interacting agents.
  pp. 2165--2174 (07 2017). \doi{10.1109/CVPR.2017.233}

\bibitem{lee2017desire}
Lee, N., Choi, W., Vernaza, P., Choy, C.B., Torr, P.H., Chandraker, M.: Desire:
  Distant future prediction in dynamic scenes with interacting agents. In:
  Proceedings of the IEEE Conference on Computer Vision and Pattern
  Recognition. pp. 336--345 (2017)

\bibitem{Conditional}
Li, J., Ma, H., Tomizuka, M.: Conditional generative neural system for
  probabilistic trajectory prediction. 2019 IEEE/RSJ International Conference
  on Intelligent Robots and Systems (IROS) pp. 6150--6156 (2019)

\bibitem{social-wagdat}
Li, J., Ma, H., Zhang, Z., Tomizuka, M.: Social-wagdat: Interaction-aware
  trajectory prediction via wasserstein graph double-attention network. arXiv
  preprint arXiv:2002.06241  (2020)

\bibitem{liu2020social}
Liu, Y., Yan, Q., Alahi, A.: Social nce: Contrastive learning of socially-aware
  motion representations. arXiv preprint arXiv:2012.11717  (2020)

\bibitem{makansi2019overcoming}
Makansi, O., Ilg, E., Cicek, O., Brox, T.: Overcoming limitations of mixture
  density networks: A sampling and fitting framework for multimodal future
  prediction. In: Proceedings of the IEEE Conference on Computer Vision and
  Pattern Recognition. pp. 7144--7153 (2019)

\bibitem{uber-scene}
Niedoba, M., Cui, H., Luo, K., Hegde, D., Chou, F.C., Djuric, N.: Improving
  movement prediction of traffic actors using off-road loss and bias
  mitigation. Machine Learning for Autonomous Driving Workshop at the 33rd
  Conference on Neural Information Processing Systems (NeurIPS 2019)

\bibitem{pytorch}
Paszke, A., Gross, S., Chintala, S., Chanan, G., Yang, E., DeVito, Z., Lin, Z.,
  Desmaison, A., Antiga, L., Lerer, A.: Automatic differentiation in pytorch
  (2017 NiPS Talk)

\bibitem{weakly_supervised}
Pathak, D., Krahenbuhl, P., Darrell, T.: Constrained convolutional neural
  networks for weakly supervised segmentation. In: Proceedings of the IEEE
  international conference on computer vision. pp. 1796--1804 (2015)

\bibitem{survey_knowledge}
von Rueden, L., Mayer, S., Beckh, K., Georgiev, B., Giesselbach, S., Heese, R.,
  Kirsch, B., Pfrommer, J., Pick, A., Ramamurthy, R., et~al.: Informed machine
  learning--a taxonomy and survey of integrating knowledge into learning
  systems. arXiv preprint arXiv:1903.12394  (2020)

\bibitem{wta}
Rupprecht, C., Laina, I., DiPietro, R., Baust, M., Tombari, F., Navab, N.,
  Hager, G.D.: Learning in an uncertain world: Representing ambiguity through
  multiple hypotheses. In: Proceedings of the IEEE International Conference on
  Computer Vision. pp. 3591--3600 (2017)

\bibitem{carnet}
Sadeghian, A., Legros, F., Voisin, M., Vesel, R., Alahi, A., Savarese, S.:
  Car-net: Clairvoyant attentive recurrent network. In: Ferrari, V., Hebert,
  M., Sminchisescu, C., Weiss, Y. (eds.) Computer Vision -- ECCV 2018. pp.
  162--180. Springer International Publishing, Cham (2018)

\bibitem{the-simpler-the-better}
Sch{\"o}ller, C., Aravantinos, V., Lay, F., Knoll, A.: What the constant
  velocity model can teach us about pedestrian motion prediction. IEEE Robotics
  and Automation Letters  \textbf{5},  1696--1703 (2020)

\bibitem{silver2018residual}
Silver, T., Allen, K., Tenenbaum, J., Kaelbling, L.: Residual policy learning.
  arXiv preprint arXiv:1812.06298  (2018)

\bibitem{mult_future_pred}
Tang, C., Salakhutdinov, R.R.: Multiple futures prediction. In: Advances in
  Neural Information Processing Systems. pp. 15398--15408 (2019)

\bibitem{tang2016estimating}
Tang, J., Song, Y., Miller, H.J., Zhou, X.: Estimating the most likely
  space--time paths, dwell times and path uncertainties from vehicle trajectory
  data: A time geographic method. Transportation Research Part C: Emerging
  Technologies  \textbf{66},  176--194 (2016)

\bibitem{leader-follower}
Treiber, M., Hennecke, A., Helbing, D.: Congested traffic states in empirical
  observations and microscopic simulations. Physical review E  \textbf{62}(2),
  ~1805 (2000)

\bibitem{social-att}
Vemula, A., Muelling, K., Oh, J.: Social attention: Modeling attention in human
  crowds. In: 2018 IEEE international Conference on Robotics and Automation
  (ICRA). pp.~1--7. IEEE (2018)

\bibitem{willard2020integrating}
Willard, J., Jia, X., Xu, S., Steinbach, M., Kumar, V.: Integrating
  physics-based modeling with machine learning: A survey. arXiv preprint
  arXiv:2003.04919  (2020)

\bibitem{xie2019data}
Xie, D.F., Fang, Z.Z., Jia, B., He, Z.: A data-driven lane-changing model based
  on deep learning. Transportation research part C: emerging technologies
  \textbf{106},  41--60 (2019)

\bibitem{xu2015asymmetric}
Xu, X., Pang, J., Monterola, C.: Asymmetric optimal-velocity car-following
  model. Physica A: Statistical Mechanics and its Applications  \textbf{436},
  565--571 (2015)

\bibitem{zeng2020tossingbot}
Zeng, A., Song, S., Lee, J., Rodriguez, A., Funkhouser, T.: Tossingbot:
  Learning to throw arbitrary objects with residual physics. IEEE Transactions
  on Robotics  (2020)

\bibitem{interactiondataset}
Zhan, W., Sun, L., Wang, D., Shi, H., Clausse, A., Naumann, M., K\"ummerle, J.,
  K\"onigshof, H., Stiller, C., de~La~Fortelle, A., Tomizuka, M.: {INTERACTION}
  {Dataset}: {An} {INTERnational}, {Adversarial} and {Cooperative} {moTION}
  {Dataset} in {Interactive} {Driving} {Scenarios} with {Semantic} {Maps}.
  arXiv:1910.03088 [cs, eess]  (2019)

\bibitem{zhang2019simultaneous}
Zhang, X., Sun, J., Qi, X., Sun, J.: Simultaneous modeling of car-following and
  lane-changing behaviors using deep learning. Transportation research part C:
  emerging technologies  \textbf{104},  287--304 (2019)

\bibitem{benz}
Ziegler, J., Bender, P., Schreiber, M., Lategahn, H., Strauss, T., Stiller, C.,
  Dang, T., Franke, U., Appenrodt, N., Keller, C.G., et~al.: Making bertha
  drive—an autonomous journey on a historic route. IEEE Intelligent
  transportation systems magazine  \textbf{6}(2),  8--20 (2014)

\end{thebibliography}
\clearpage

\end{document}